\documentclass{article}
\usepackage{spconf,amsmath,graphicx}
\usepackage{booktabs}

\usepackage{times}
\usepackage{xcolor}
\usepackage{soul}
\usepackage[utf8]{inputenc}
\usepackage{url}

\usepackage[small]{caption}

\usepackage{amssymb}
\usepackage{times}
\usepackage{epsfig}
\usepackage{graphicx}
\usepackage{amsthm}
\usepackage{amsmath}
\usepackage{amssymb}
\usepackage{booktabs}
\usepackage{subcaption}
\usepackage{algorithm,algorithmic}
\usepackage{color}
\usepackage{caption}
\usepackage{adjustbox}
\usepackage{multirow}
\usepackage{array}
\usepackage{pdfpages}
\usepackage{siunitx}
\usepackage{graphicx}

\def\ie{\textit{i.e.,~}}

\def\etal{\textit{et al.}~}

\newcommand{\rsec}[1]{Section~\ref{#1}}

\newcommand{\rfig}[1]{Figure~\ref{#1}}
\newcommand{\rtab}[1]{Table~\ref{#1}}

\title{Deep inspection: an electrical distribution pole parts study via deep neural networks}
\name{Liangchen Liu$^1$, Teng Zhang$^1$, Kun Zhao$^1$, Arnold Wiliem$^1$, Kieren Astin-Walmsley$^2$, Brian Lovell$^1$}
\address{$^1$The University of Queensland, $^2$Energy Queensland, Australia}
\begin{document}
%
\maketitle
\begin{abstract}
Electrical distribution poles are important assets in electricity supply.
These poles need to be maintained in good condition to ensure they protect community safety, maintain reliability of supply, and meet legislative obligations.
However, maintaining such a large volumes of assets is an expensive and challenging task. 
To address this, recent approaches utilise imagery data captured from helicopter and/or drone inspections.
Whilst reducing the cost for manual inspection, manual analysis on each image is still required. 
As such, several image-based automated inspection systems have been proposed.
In this paper, we target two major challenges: tiny object detection and extremely imbalanced datasets, which currently hinder the wide deployment of the automatic inspection. 
We propose a novel two-stage zoom-in detection method to gradually focus on the object of interest. 
To address the imbalanced dataset problem, we propose the resampling as well as reweighting schemes to iteratively adapt the model to the large intra-class variation of major class and balance the contributions to the loss from each class.
Finally, we integrate these components together and devise a novel automatic inspection framework. 
Extensive experiments demonstrate that our proposed approaches are effective and can boost the performance compared to the baseline methods.
\end{abstract}

\begin{keywords}
electrical distribution pole, integrated inspection system, deep neural networks, object detection, imbalanced data classification 
\end{keywords}
\section{Introduction}
Electrical distribution poles are critical infrastructure for the modern world as they help to supply electricity into local communities.
As the integration of the renewable energy sources and increasingly complication of modern power system~\cite{li2018integrating}, the electrical distribution devices and whole power delivery system face unprecedented challenges to guarantee the transmission security.
As such, it is important to perform regular monitoring and maintenance on the poles. 
One recent method~\cite{6995104} to perform this is by manually analysing from imagery data produced from helicopter and/or drone inspections. 
However, as more data is generated, this manual process becomes time consuming, labour intensive and expensive. 

A distribution pole condition is generally described as a textual description highlighting potential issue/condition of an individual pole component (\ie grade 4 corrosion in the first cross arm). 
Careful inspection of each individual component is time consuming.
In this work, we tackle this problem by utilising and adapting recent methods in deep neural networks~\cite{simonyan2014very, RN224}.
Specifically, we confine ourselves on pole cap missing condition. 
Pole cap missing condition is chosen as it poses two problems of which are interest to the community: (1) pole cap size is significantly smaller than the other objects; and (2) rare occurrences of pole cap missing condition.

Several recent works have shown the issue in the current state-of-the-art deep neural network methods in handling small objects~\cite{redmon2017yolo9000, hu2017finding, wu2011numerical}. 
For instance, Hu and Ramanan~\cite{hu2017finding} train separate detectors for different scales to detect small faces. 
In this work, we propose a two-stage zoom-in detector that will first detect the distribution pole and then detect the pole cap by focusing at the most likely location of pole cap in the cropped distribution pole image.

Once the pole cap has been identified, the next step is to train a binary classifier which detects the condition of the pole cap. 
As there is a highly imbalanced distribution between the normal cases (\ie pole cap exists) and abnormal cases (\ie pole cap missing), we propose to perform resampling and reweighing approach.

\begin{figure}[tp]
\centering
    \includegraphics[width=1\linewidth]{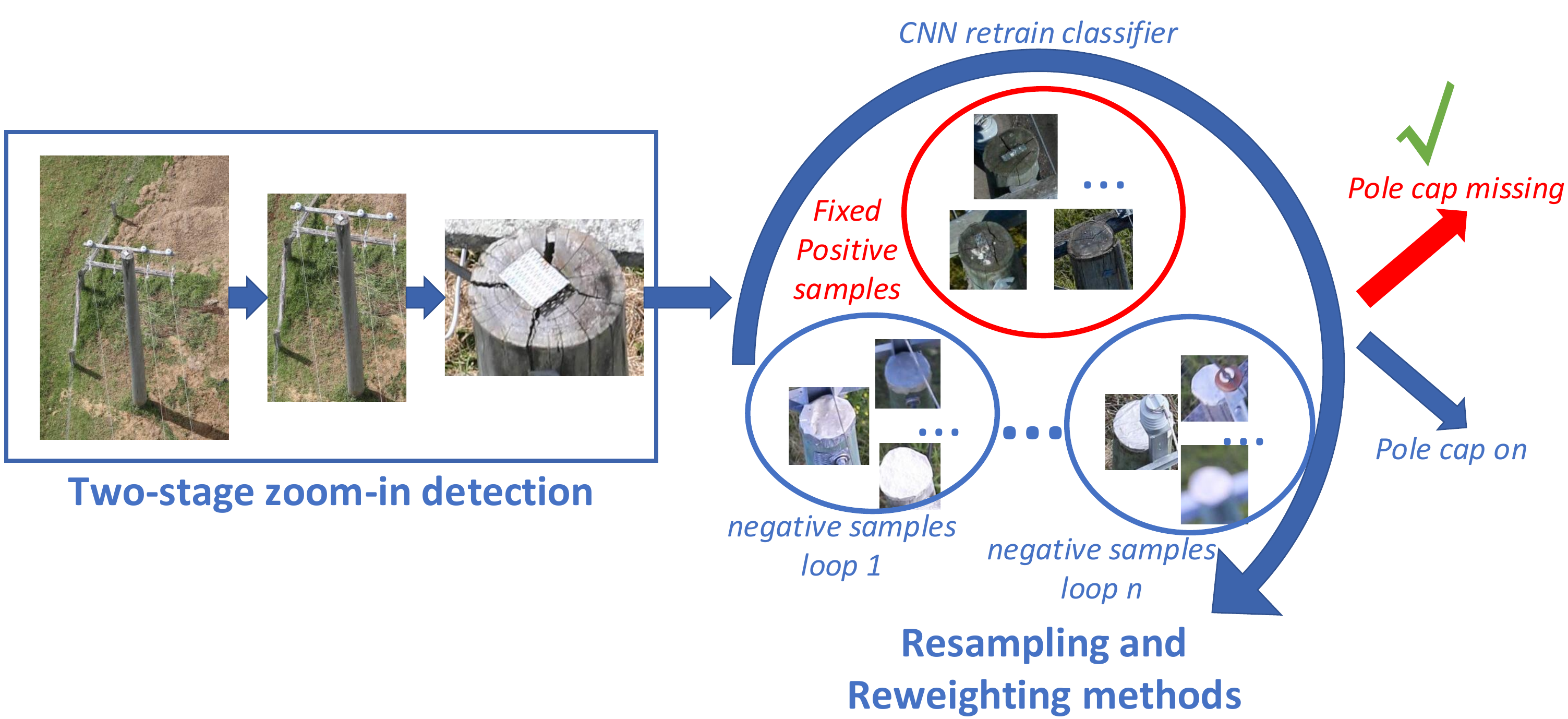}
    \caption{The illustration of the proposed electrical distribution pole inspection method}
\label{fig:idea}
\end{figure}

The main idea of our paper is illustrated in Figure~\ref{fig:idea}. 
Our contributions can be summarized as follows.

\begin{enumerate}
\item we propose an automatic pole inspection system that will first locate the pole component and then perform automated condition analysis on the component
\item we propose the two-stage zoom-in detection method as the first component in the system to solve the tiny object detection problem
\item we propose the negative sample resampling and reweighting approaches to target the large with-in class variance and lack of positive samples
\item we provide results, analysis from experiments.
These results will provide useful insights for tackling problems in performing fully automated distribution tower condition analysis
\end{enumerate}

%
%
%
%
%

\section{Related works}
We first discuss the recent works in distribution pole inspection. 
Then related works in the deep neural network will be discussed.
\subsection{Electrical distribution pole inspection}
Many distribution pole inspections are still performed by the inspection team travelling on foot~\cite{5345712}. 
They use varied methods such as visible spectrum cameras, infrared cameras, binoculars, and corona detection camera to perform the visual inspection on the pole and attached components such as conductors, transformers and insulators.
This type of inspection is labour intensive and time consuming. 
Extreme weather and difficult terrains can also present challenges.

To that end, helicopter inspection can be utilised~\cite{6820566}. 
The inspection can then be performed online or offline.
The image size produced by this type of inspection is usually very large compared to the objects of interest such as tower components. 
This leads to the tiny object detection problem.
Note that, the data we used in this paper was produced from this type of inspection.

Various individual tasks are performed in the inspection of the distribution pole targeted on different parts of the whole pole~\cite{nguyen2018automatic}. 
They can be inspected online by the inspectors and on-site inference system or offline by the image analysis tools~\cite{LI20081}.
In this paper, we mainly focus on the pole cap missing problem and pole type classification, which will help determine the type of faults on the body of the pole.
 
\subsection{Deep neural network}
There has been a big leap forward on the performance of vision-based applications due to the deep learning methods~\cite{RN224, simonyan2014very, krizhevsky2012imagenet, wu2018deep}.
In this work, we primarily aim at tiny object detection problem and training deep neural network classifier with imbalanced data.

\noindent
\textbf{Object detection -- }We utilise Faster R-CNN~\cite{RN224} to address the pole component localization task.
It is one of the state-of-the-art detection methods which has been proven to be effective for various real-world detection tasks. Faster R-CNN introduces the Region Proposal Network (RPN) that shares full-image convolutional features with the detection network, thus enabling nearly cost-free region proposals. An RPN is a fully-convolutional network that simultaneously predicts the object bounding box and objectness score at each position. With a simple alternating optimization, RPN and Fast R-CNN~\cite{girshick2015fast} can be trained to share convolutional features.

As to the backbone deep neural network, we use VGG16~\cite{simonyan2014very} for both the classification and detection. 
VGG16 improves from AlexNet~\cite{krizhevsky2012imagenet} by replacing large filters with small $3\times3$ filters. 
It has been shown that stacked smaller size kernel filters are better than filters with larger size kernel.   
The VGG16 net achieves the top-5 accuracy of 92.3\% on ImageNet.

\noindent
\textbf{Dealing with imbalanced data -- }Resampling networks~\cite{Shrivastava_2016_CVPR, Ahmed_2015_CVPR, GHAZIKHANI2013535} on the hard samples is an effective method to improve the performance.
Shrivastava~\etal~\cite{Shrivastava_2016_CVPR} proposed to use an automatic selection of hard examples and iteratively make training more effective.
Ahmed~\etal~\cite{Ahmed_2015_CVPR} resampled the fully connected (top) layer of the network using a set containing as many of the difficult negative sample pairs as positive sample pairs to improve the performance.
Ghazikhani~\etal~\cite{GHAZIKHANI2013535} introduce the general resampling and reweighting usage in imbalanced data stream problem.

Reweight~\cite{eigen2015predicting, badrinarayanan2015segnet} is a technique used for rebalancing the class distribution to solve the imbalanced dataset problem in deep neural network.
It is often used in the image segmentation problem where classes with small numbers of pixels appear.
Its main idea is to modify the weight of the loss function for each sample from each class to balance the contribution to the loss from each class.  

\subsection{Current deep neural network based pole inspection methods review}
Although considerable improvement~\cite{6995104, 6693261, 7502544, watanabe2018electric, zhang2018using} has been performed in the automation of the distribution pole inspection,
there are only few a works utilising the recent deep neural networks~\cite{10.1007/978-3-319-59126-1_21, RN381}.
Nordeng~\etal~\cite{10.1007/978-3-319-59126-1_21} and Nguyena~\etal~\cite{RN381} both propose to use Faster R-CNN to develop the component detection system. 
In contrast to these works, we use the prior of the top half image of sample to perform the two-stage zoom-in detection which is able to increase the performance.
The two major challenges of tiny object detection and extremely imbalanced dataset are also discussed in~\cite{RN381}.
However, in this work, we propose an integrated inspection system to solve them in a complete pipeline.

\section{Proposed Framework}
We first introduce the two-stage zoom-in detection approach to handle objects of tiny size in a large image. 
Also, to deal with the problem of lacking positive samples, we propose iteratively resampling and reweighting approaches to maintain data balance during each re-training by replacing negative samples only.

\subsection{The two-stage zoom-in detection}
\label{zoom-in}
The two-stage zoom-in detection method is simple yet effective when detecting small objects. 
The main idea of this method is to reuse the current existing deep learning method to detect and crop out the important sub-region of the original large image with a coarse detector before detecting the target tiny object in the cropped sub-region with the second detector.

In the scenario of the pole cap assessment task, the proposed two-stage zoom-in detection approach first detects and crops the whole pole and searches for the cap in the cropped smaller image more effectively. 
Instead of a single stage detection approach operating on the original large image, our method performs pole cap detection in a much smaller image.
Therefore, the pole cap remains "visible" in a reasonable size when feeding into the network. 
We use the state-of-the-art Faster R-CNN~\cite{RN224} methods to first detect the whole pole object in the original image.
Then, after the region of the whole pole is detected, we reuse a second Faster R-CNN model to detect the pole cap. 

\begin{figure}[htbp]
\centering
    \includegraphics[width=0.9\linewidth]{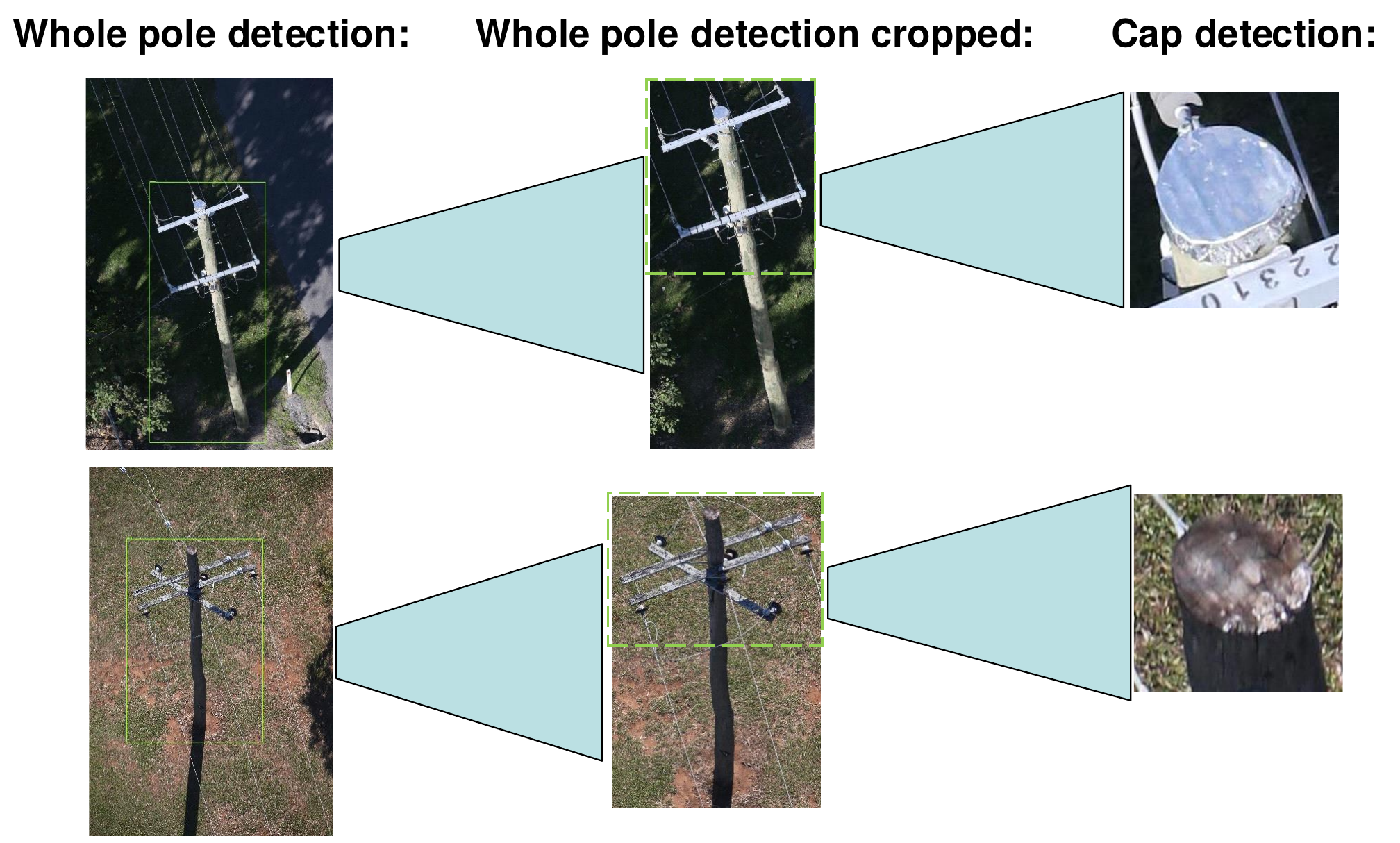}
    \caption{The illustration of the two-stage zoom-in detection}
\label{fig:zoom}
\end{figure}

The illustration of the two-stage zoom-in detection is shown in~\rfig{fig:zoom}.
As we can see the green bounding box of the whole pole is detected and cropped using the first Faster R-CNN model.
Then, the sub-region is cropped again by half to get the top part where contains the cross-arms and pole cap. 
This can be done with simple entropy/edges comparison as the pole base area is sparse. 
Finally, we use our second pole cap detector to locate the cap within the upper half of the whole pole region.

\subsection{The negative sample resampling and reweighting approach}
In well maintained assets, the probability of abnormal condition is assumed to be very low. 
This affects the data distribution which is significantly skewed towards the negative exemplars.
As such, we propose a data resampling approach which is also inspired by~\cite{Shrivastava_2016_CVPR}.
First, we sample the same amount of negative training samples as the positive samples from the large negative sample pool. 
Also due to the lack of positive data, we use transfer learning method to deal with this task rather than training the model from scratch.
In this task, we use the VGG-16 model pretrained from the large-scale ImageNet dataset. 
However, this pretrained model is not sufficient to address the data imbalance problem. 
Thus, we propose the resampling method to target the skew dataset problem.
The resampling method performs several iterations of the training process by fixing the small positive set and the same balanced amount of negative from the sample pool. 
The benefit of the resampling method lies in that the model can be trained to be adaptive to the large intra-class variation with a large number of both negative and positive samples. 
Meanwhile, the method can still keep the classification capability of the model for the small sample side. 

Furthermore, we also propose a reweighting scheme to target the problem.
The key idea of reweighting lies in balancing the contributions from each class to the objective function. 
We multiply a reweighting coefficient on the loss function of each sample with respect to its class to offset the contribution from the minor class which has a much lesser frequency.
By this way, the reweighting method can divert the attention of classifier to the class while training.
Therefore, the classifier gets larger penalty when misclassifying a sample from minor class than major class.
For example, when performing the classification task on 10 positive samples vs 1000 negative samples, the reweighting scheme could make the classifier penalized 100 times larger in misclassifying one positive sample than in one negative sample.


\section{Experiments}
In this part, we will introduce four experiments and their different settings: pole cap detection, pole cap missing and wood pole type classification with proposed resampling, and pole cap classification with the proposed reweighting method.
\subsection{Experiment setting}
We used a dataset comprising 103,649 entries, each of which is an individual sample with its description. 
However, there are no detection labels provided. 
Therefore, we manually labelled various objects including pole cap, cross arm, pole, whole pole for 428 image samples.
We split the 428 image samples into training dataset with 216 samples and testing dataset with 212 samples to perform the pole cap detection experiment.
We finetune the Faster R-CNN model with VGG16 network as backbone pretrained from the COCO dataset for our detection task and finetune the VGG16 network pretrained from ImageNet dataset for our classification task.
We perform 20,000 epochs on all the tasks and use Average Precision for detection task and ROC curve and AUC metrics for classification task.
\subsection{Results and analysis}
\textbf{Zoom-in detection:} We perform two-stage zoom-in detection as discussed in~\rsec{zoom-in} and compare it with the pole cap detection directly from the original image.
We use the primary challenge metric Average Precision (AP) according to the protocol in~\cite{coco, lin2014microsoft} to evaluate the results.
The results are shown in~\rtab{tab:map}.

\begin{table}[htp]
\centering
\caption{Results of two-stage zoom-in detection}
\begin{tabular}{ccccc}
\cline{1-3}
    & baseline detection & two-stage zoom-in detection &  &  \\ \cline{1-3}
mAP &        0.507        &             0.796            &  &  \\ \cline{1-3}
\end{tabular}
\label{tab:map}%
\end{table}

As in the results, our two-stage zoom-in detection largely outperforms the baseline detection method, which demonstrates the efficacy of our method.

\begin{figure}[ht!]
	\centering
	\begin{subfigure}[t]{0.23\textwidth}
		\includegraphics[width=\linewidth]{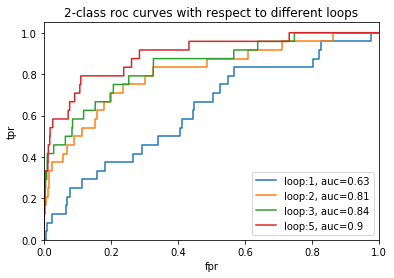}
	\caption{pole cap results}
	\label{fig:retrain} 
	\end{subfigure}
	\begin{subfigure}[t]{0.23\textwidth}
		\includegraphics[width=\linewidth]{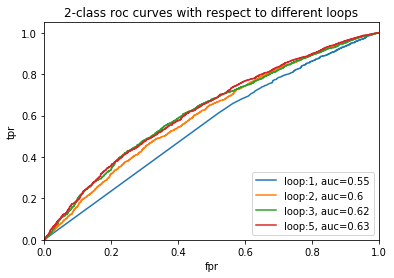}
	\caption{pole type results}
	\label{fig:pole_type} 
	\end{subfigure}
	\caption{The results of the negative sample resampling approach for pole cap and pole type}
	\label{fig:retrain_all}
\end{figure}

\noindent\textbf{Resampling approach:} From our observation, the dataset for the pole cap is extremely imbalanced.
There are only 110 available positive samples compared to 103,561 negative samples.
We first perform the two-stage zoom-in detection and crop the pole cap part for each image in the dataset and crop the respective pole cap areas.
To make the most of our current data, we set 88 positive training data (80\% of all positive samples) and 22 positive testing data (20\% of all positive samples), then we divide all the negative data in half and set one half as negative training data pool (where we will conduct the resampling loops) and negative testing data.
Then, we iteratively sample the same amount of negative training data as the positive training data from the negative training data pool to continue retraining the classifier.
After each training loop, we perform testing on the testing set.
To better assess the performance of the pole cap missing classification, we introduce the Receiver Operating Characteristic (ROC) curve and Area Under the Curve (AUC) metrics~\cite{roc} to evaluate the performance, since the AP is mostly used for detection task and is no longer discriminative metric for classification in extremely imbalanced dataset.
the results are shown in~\rfig{fig:retrain}.

As we can see, our method is effective since AUC significantly improves when more iterations are performed (we show 1,2,3 and 5 loops). 

Similarly, we perform the same resampling experiment on the wood pole type classification, since this is also a case of extremely imbalanced examples in our dataset according to the statistics in~\rtab{tab:stat}. 
\begin{table}[htbp]
  \centering
  \small
  \caption{The statistics of pole type from our sample dataset}
    \begin{tabular}{cc}
    \toprule
    Pole\_Type & \multicolumn{1}{c}{Counts} \\
    \midrule
    Aluminium & 1 \\
    Concrete & 1027 \\
    Nailed & 2558 \\
    Rebutted & 671 \\
    Steel & 10 \\
    Unknown & 48 \\
    Wood  & 99330 \\
    Wood Multi/Trans & 2 \\
    Wood x2 - PT & 2 \\
    \bottomrule
    \end{tabular}%
  \label{tab:stat}%
\end{table}%

In this experiment, we consider the wood pole type as negative samples and all other pole types as the positive samples and have the same configuration of the experiment as in pole cap above.
The results are shown in~\rfig{fig:pole_type}.
As we can see, the results of the negative sample resampling approach for wood pole type also show decent performance. 
The resampling approach improves the AUC score within several loops similar to the pole cap results above.
However, the results are not as good as the case of the pole cap. 
We suspect that this is because of large intraclass variation in the negative samples and too few resampling iteration loops.

\noindent\textbf{Reweighting approach:} For the reweighting scheme experiment, we gradually increase the imbalance of the training dataset of the pole cap from the balanced case to 12 times more negative samples than positive samples.
We use AUC metric and also test on the testing dataset with number of samples equal to 20\% of the training set.
The results are shown in \rtab{tab:reweight}.
\begin{table}[htbp]
  \centering
  \small
  \caption{The results of the reweighting scheme}
  \begin{tabular}{ccccc}
    \toprule
          & 1 (balanced) & 3     & 6     & 12 \\
    \midrule
    Baseline & 0.941 & 0.912 & 0.853 & 0.991 \\
    Reweighting & 0.956 & 0.951 & 0.954 & 0.992 \\
    \bottomrule
    \end{tabular}%
  \label{tab:reweight}%
\end{table}%

As we can see, the results of our reweighting method are consistently better than the baseline. 
This demonstrates the efficacy of our method.

\section{Conclusion}
In this paper, we focus on the automatic pole inspection task. 
The tiny object detection and extremely imbalanced dataset are two of the main challenges impeding the practical usage of the automatic inspection. 
Three novel approaches are proposed to confront those challenges: two-stage zoom-in detection, resampling, and reweighting. 
The two-stage zoom-in detection method is able to focus on the small pole cap object gradually from the whole image through detecting the whole pole intermediately.
The resampling and reweighting schemes can iteratively adapt the model to the large intra-class variation of major class and balance the contributions to the loss from each class.
Several experiments show the efficacy of our methods.
In future work, we plan to jointly fuse the resampling and reweighting approaches together, as well as consider multi-modal data~\cite{liu2018multi} to further improve the performance of the framework. Automatic keywords or report generation~\cite{liu2017best} functions for the inspection could also be developed from this framework.

\bibliographystyle{IEEEbib}
\bibliography{strings,refs}

\end{document}